\documentclass{article}

\usepackage[main, final]{neurips_2026}

\usepackage[utf8]{inputenc} 
\usepackage[T1]{fontenc}    
\usepackage{microtype}
\usepackage{graphicx}
\usepackage{subcaption}
\usepackage{booktabs} 
\usepackage{float}
\usepackage{amsmath}
\usepackage{amsfonts}
\usepackage{nicefrac}
\usepackage{url}
\usepackage{hyperref}
\usepackage{multirow}
\usepackage[table]{xcolor}
\usepackage{wrapfig}

\newcommand{\markc}[1]{\textcolor{red}}
\definecolor{highlight}{RGB}{220, 230, 255}
\definecolor{grey}{RGB}{220, 220, 220}

\usepackage{amssymb}
\usepackage{mathtools}
\usepackage{amsthm}
\usepackage{algorithm}
\usepackage{algorithmic}

\usepackage[capitalize,noabbrev]{cleveref}

\theoremstyle{plain}
\newtheorem{theorem}{Theorem}[section]

\newtheorem{corollary}[theorem]{Corollary}
\theoremstyle{definition}

\theoremstyle{remark}
\newtheorem{remark}[theorem]{Remark}

\usepackage[disable,textsize=tiny]{todonotes}
\usepackage{bbm}

\newcommand{\secref}[1]{Sec.~\ref{#1}}

\title{Dynamic Expert Sharing: Decoupling Memory from Parallelism in Mixture-of-Experts Diffusion LLMs}

\author{%
  Hao (Mark) Chen$^{1}$, Zhiwen Mo$^{1}$, Royson Lee$^{2}$, Qianzhou Wang$^{1}$, Da Li$^{2}$, \\
  \textbf{Shell Xu Hu$^{2}$, Wayne Luk$^{1}$, Timothy Hospedales$^{2}$, Hongxiang Fan$^{1}$} \\
  $^{1}$Imperial College London, London, UK \quad $^{2}$Samsung AI Center, Cambridge, UK \\
}

\begin{document}

\maketitle

\begin{abstract}
Among parallel decoding paradigms, diffusion large language models (dLLMs) have emerged as a promising candidate that balances generation quality and throughput.
However, 
unlike traditional autoregressive (AR) decoding, this paper identifies a critical bottleneck in dLLMs with Mixture-of-Experts (MoE) architectures: ``\textbf{expert explosion}''. As the number of parallel tokens generated in dLLMs increases, the number of distinct experts activated grows nearly linearly.
This results in substantial memory traffic that pushes inference into a memory-bound regime, negating the efficiency gains and scalability of both MoE and parallel decoding. To address this challenge, we propose \textbf{Dynamic Expert Sharing (DES)}, a novel technique
that shifts MoE optimization from conventional token-centric pruning and expert skipping to sequence-level expert sharing.
To maximize expert reuse, DES identifies a compact, high-utility set of experts for an entire parallel decoding block and further improves system efficiency with custom GPU kernels.
Extensive experiments on MoE dLLMs demonstrate that, on LLaDA2.0-Mini, DES reduces unique expert activations by up to 55\% and latency by up to 38\%, while retaining up to 99\% of vanilla accuracy, effectively decoupling memory overhead from the degree of parallelism.
Code is available at \url{https://github.com/hmarkc/DES}.
\end{abstract}

\section{Introduction}
\label{sec:intro}


Recent advances in large-scale foundation models have seen the emergence of native parallel decoding paradigms, most notably diffusion large language models (dLLMs)~\cite{zhu2025llada, ye2025dream}. The state-of-the-art dLLMs typically leverage block-based parallel decoding~\cite{arriola2025block, wu2025fast} to achieve performance parity with traditional autoregressive~(AR) models through concurrent token generation. 
To scale these architectures effectively, dLLMs increasingly integrate Mixture-of-Experts (MoE)~\cite{guo2025deepseek, yang2025qwen3, agarwal2025gpt} as a core architectural component~\cite{bie2025llada2, zhu2025llada, cheng2025sdar}. The primary advantage of MoE lies in its ability to decouple the model's overall knowledge capacity from its per-token computational cost. By activating only a sparse subset of experts for each input, MoE allows dLLMs to scale to massive parameter counts, significantly enhancing representational power without a proportional increase in FLOPs.

\begin{figure}[t]
    \centering
    \includegraphics[width=0.99\linewidth]{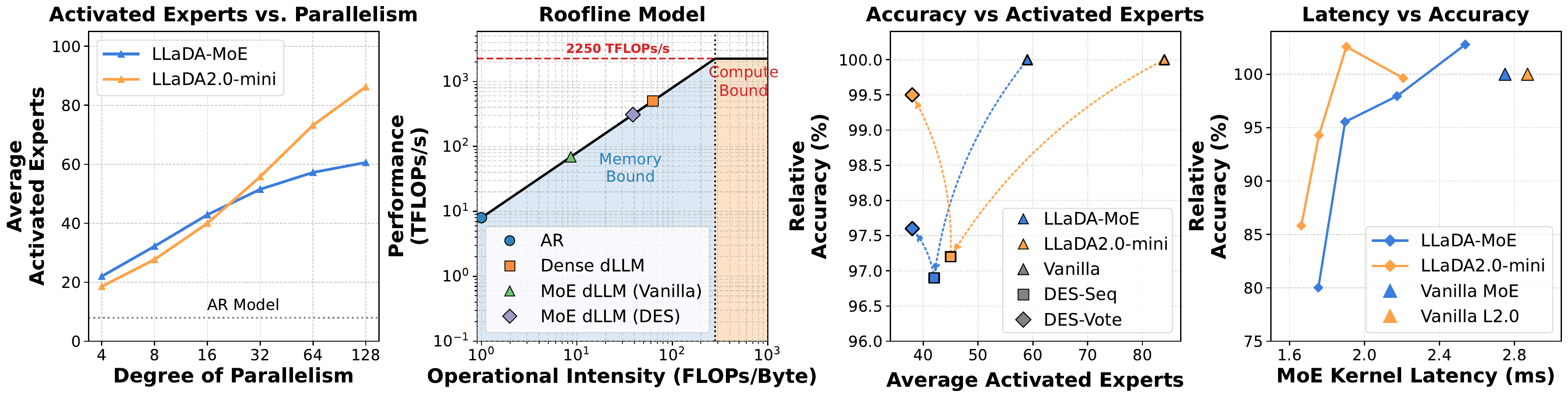}
     \caption{From left to right: (a) Growth of unique activated experts vs.\ block size. (b) Roofline model showing MoE dLLMs are more memory-bound than dense counterparts and how DES mitigates this bottleneck. (c) Activated experts vs.\ relative accuracy, where DES-Vote extends the Pareto frontier. (d) MoE kernel latency vs.\ relative accuracy. Accuracy in (c, d) is relative to the vanilla baseline. \textbf{Setup}: single B200, batch size 1, generation length 2048; block size 32 except (a) sweeps 4--128. (a--c) average over HumanEval, MBPP, MATH500, GSM8K; (d) uses MBPP.
     \vspace{-5mm}
     }
    \label{fig:intro_overview}
\end{figure}

The integration of MoE into dLLMs, however, introduces an efficiency paradox, illustrated by our roofline analysis (Figure~\ref{fig:intro_overview}b). While transitioning from AR to parallel dLLM increases operational intensity, the sparse activation of MoE dLLMs makes them more memory-bound than their dense counterparts. 
In these memory-bound scenarios, typical of multi-GPU and CPU-offloading environments~\cite{cao2025moe, yu2025efficient}, latency is dominated by the traffic from High Bandwidth Memory (HBM) to Static RAM (SRAM) required to load unique expert weights. 
This bottleneck is intensified by an ``\textbf{expert explosion}'' inherent to parallel generation: whereas AR decoding processes a single token per step, dLLMs process $N$ tokens simultaneously (Figure~\ref{fig:intro_overview}a). 
Consequently, if experts are selected independently, the unique expert load scales nearly linearly with the block size $N$, rapidly exceeding the steady-state memory traffic of AR decoding (Figure~\ref{fig:intro_overview}b). 
Crucially, as the parallel block scales, the overhead of transferring divergent experts from HBM to on-chip memory can negate the compute savings of MoE, potentially rendering these models slower than their dense equivalents. 
This overhead imposes a systemic ceiling on the throughput gains promised by parallelization.

Existing dynamic efficiency optimizations for MoE, such as expert skipping~\cite{lu2024not, huang2024mixture, aghdam2024moe}, are primarily designed for AR models and remain strictly token-centric. While effective at reducing per-token computational FLOPs, these methods prioritize local expert sparsity rather than exploiting cross-token semantic coherence to reduce expert redundancy. Consequently, they fail to mitigate the global HBM traffic bottleneck, as the unique expert weights loaded per layer are not optimized at the sequence level. Crucially, these approaches do not directly address the memory traffic overhead encountered in block-wise parallel decoding, where the collective expert footprint, rather than per-token compute, becomes the primary constraint on throughput.

To address these challenges, we propose \textbf{Dynamic Expert Sharing (DES)}, a novel method designed to minimize the expert footprint by forming consensus across parallel tokens. 
Our approach shifts the optimization objective from per-token pruning to sequence-level sharing. 
We hypothesize that, unlike standard AR batching where independent tokens may activate disparate experts, \textit{parallel decoding in dLLMs inherently involves semantically coupled tokens that exhibit significant overlap in expert demand}. 
To exploit this, we introduce \textbf{coreset selection}, where a minimal, high-utility set of experts is dynamically identified at runtime to serve the entire parallel block. We explore two specific strategies:
\textbf{(1) Intra-Sequence Sharing (DES-Seq):} We take the union of experts chosen independently by each token in the block. This allows for vectorized execution, eliminating the need to handle divergent expert assignments for individual positions. \textbf{(2) Saliency-Aware Voting (DES-Vote):} A novel mechanism where tokens collectively vote for experts based on their weighted router saliency. This prioritizes the most critical experts by focusing on the holistic semantic complexity of the sequence.

By restricting expert selection to a shared coreset, we significantly reduce HBM-to-SRAM traffic, effectively mitigating the memory traffic overhead of parallel MoE decoding at a small accuracy cost.
As shown in Figure~\ref{fig:intro_overview}c and Figure~\ref{fig:intro_overview}d, DES reduces the active experts by up to 55\% and the MoE kernel latency by up to 38.0\% while achieving performance comparable to the vanilla baseline.

Our contributions are summarized as follows:
\begin{enumerate}
    \item We identify and quantify the ``expert explosion'' bottleneck in parallel MoE decoding, demonstrating how independent expert selection across parallel tokens induces a critical memory traffic overhead (Section~\ref{sec:motivation}).
    \item We propose \textbf{Dynamic Expert Sharing}, a method that shifts MoE optimization from token-centric pruning to sequence-level expert sharing, leveraging a novel saliency-aware voting mechanism to maximize expert reuse (Section~\ref{sec:method}).
    \item Through extensive benchmarking on MoE dLLMs, we demonstrate that DES achieves significant memory saving and speedup with negligible accuracy drop, effectively decoupling memory traffic from the degree of parallelism (Section~\ref{sec:exp}).
\end{enumerate}


\section{Background}

\paragraph{Parallel Decoding.}
Parallel decoding has been adopted for LLMs to reduce the number of memory-bound forward passes. 
These approaches utilize either trained tokens~\cite{chen2024hardware, lin2025bita} or $n$-grams retrieved from context~\cite{fu2024break} as inputs to an AR model, enabling the parallel generation of multiple tokens. 
However, limited backbone re-training caps generation quality, restricting these models to speculative decoding settings that require an extra verification step.

\paragraph{Diffusion Large Language Models.} 
To fully realize the acceleration potential of parallel generation, natively parallel models~\cite{gat2025set, zhu2025llada} are being trained from scratch with multi-token prediction objectives. In particular, Diffusion Large Language Models (dLLMs) have emerged as a promising candidate to balance generation quality and efficiency~\cite{zhu2025llada, ye2025dream, arriola2025block}. These models ingest multiple masked tokens and process them concurrently, enabling parallel decoding without the need for verification steps. To further enhance dLLM throughput, recent research has introduced approximate KV cache and sampling strategies~\cite{wu2025fast, wei2025accelerating, lu2025adablock} alongside specialized inference frameworks~\cite{ma2025dinfer}. 
In concurrent work, TEAM~\cite{wei2026team} accelerates MoE dLLMs by leveraging temporal-spatial consistency with adaptive decoding and speculative exploration strategies.

\paragraph{Efficient Mixture-of-Experts.}
Mixture-of-Experts (MoE) has emerged as the de facto architecture to scale up dLLMs~\cite{bie2025llada2, zhu2025llada, cheng2025sdar}. 
Formally, an MoE layer replaces the standard feedforward network (FFN) with a set of $M$ independent experts $\{E_1, \dots, E_M\}$, where each expert $E_i$ is a sub-network and $D$ is the hidden dimension. A gating or routing function $G: \mathbb{R}^D \to \mathbb{R}^M$ assigns a score to each expert for a given input hidden state $x$. The output of the MoE module is computed as the weighted sum of the $K$ highest-scoring experts:
$\text{MoE}(x) = \sum_{i \in \mathcal{S}} \frac{G(x)_i}{\sum_{j \in \mathcal{S}} G(x)_j} E_i(x)$,
where $\mathcal{S} = \text{TopK}(G(x), K)$ is the set of indices corresponding to the $K$ experts with the highest routing scores.  
For dLLMs, the total number of unique experts required across a batch of $N$ tokens is $|\bigcup_{n=1}^N \mathcal{S}_n|$, which typically grows as the degree of parallelism increases.
Existing strategies for MoE efficiency primarily follow two paradigms: offline expert pruning and dynamic expert skipping. 
Offline approaches, such as expert pruning~\cite{chen2022task, zhang2025diversifying} and merging~\cite{li2023merge}, aim to reduce the static parameter footprint through structural sparsity, quantization, or weight consolidation. In contrast, dynamic expert skipping~\cite{huang2025modes, lu2024not, huang2024mixture, aghdam2024moe} leverages runtime information to bypass redundant expert computations on-the-fly. Additionally, recent works~\cite{oncescu2025opportunistic, vankov2026xshare, mcdanel2026moe} attempted expert sharing across batches for AR decoding. 


\paragraph{Key Difference.} DES is fundamentally distinct from prior dynamic MoE optimizations along three axes: \emph{(i) sharing vs.\ skipping}: rather than activating fewer experts per token to save FLOPs, DES concentrates a shared expert pool across tokens to cut unique HBM traffic; \emph{(ii) sequence-level vs.\ token-level}: DES aggregates routing information across the entire parallel block into a collective decision instead of making isolated per-token ones; and \emph{(iii) sequence vs.\ batch dimension}: DES exploits the semantic coupling of tokens within a single block, orthogonal to batch-level sharing (e.g., OEA~\cite{oncescu2025opportunistic}) and to offline pruning. Existing online optimizations are largely tailored for AR models and do not target the memory-traffic bottleneck inherent to dLLM parallel decoding.

\begin{figure}[t]
    \centering
    \includegraphics[width=0.99\linewidth]{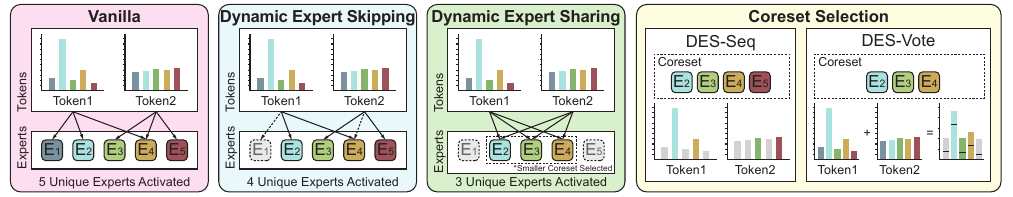}
    \caption{Overview of the Dynamic Expert Sharing method. (a) Vanilla MoE dLLM independently routes multiple tokens, leading to many unique activated experts. (b) Dynamic Expert Skipping reduces local per-token computation (indicated by dotted lines) but often fails to optimize the global unique expert load. (c) Dynamic Expert Sharing employs sequence-level coreset selection (via DES-Seq or DES-Vote) to identify high-utility experts globally, minimizing unique expert weights transferred from HBM. 
    (d) DES-Vote enforces sequence-level consensus by aggregating router weights across the parallel block, contrasting with the independent per-token selection in DES-Seq.
    Greyed out boxes and bars represent experts not selected and not fetched to on-chip memory.
    \vspace{-5mm}
    }
    \label{fig:method_overview}
\end{figure}

\section{Motivation}
\label{sec:motivation}

\begin{wrapfigure}{r}{0.5\linewidth}
    \vspace{-5mm}
    \centering
    \includegraphics[width=\linewidth]{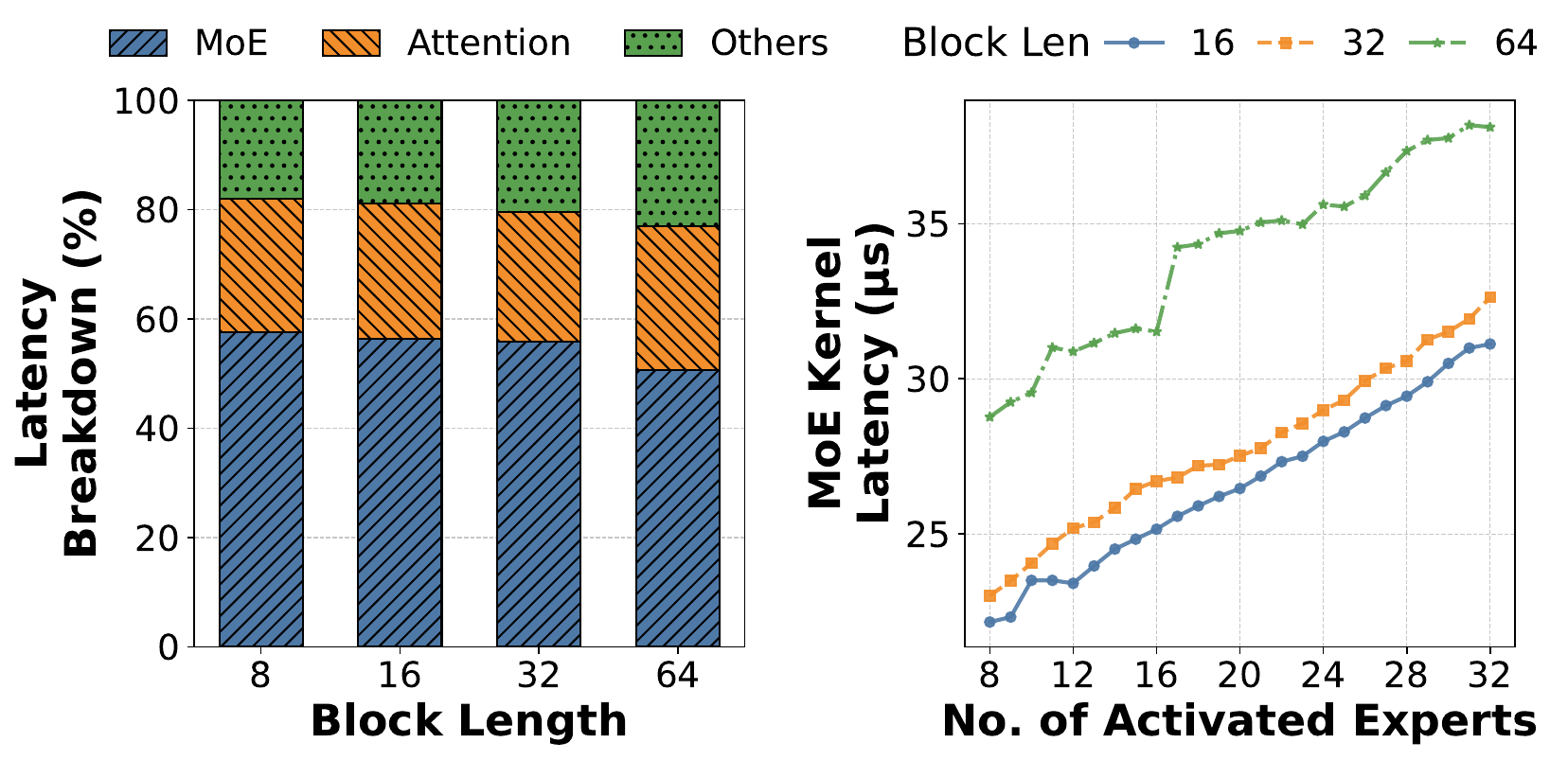}
    \caption{
     Left: Latency breakdown across MoE FFN, attention, and other components. Right: MoE kernel latency versus activated experts for different block lengths.
    \vspace{-5mm}
    }
    \label{fig:latency_breakdown}
\end{wrapfigure}

The primary motivation for parallel decoding in dLLMs is to overcome the sequential bottleneck of AR generation, which is notoriously memory-bound due to the low arithmetic intensity of processing a single token per forward pass. While dLLMs scale concurrent tokens to increase arithmetic intensity, 
the integration of MoE reintroduces the memory-bound bottleneck (Figure~\ref{fig:intro_overview}b). 
This is exacerbated by both algorithmic constraints, where smaller parallel blocks (\textit{e.g.}, 16/32 tokens) are necessary to maintain competitive accuracy \cite{arriola2025block}, and hardware trends, where the FLOPs/byte ratio of modern GPUs continues to outpace memory bandwidth \cite{ma2026challenges}. 
As a result, MoE dLLM inference remains memory-bound, with latency dominated by loading unique expert weights, a demand that grows far faster than in AR decoding~\cite{yu2025efficient}.

Within this regime, the MoE FFN layer emerges as the dominant latency contributor (Figure \ref{fig:latency_breakdown}). Following the simplified latency model from previous work \cite{oncescu2025opportunistic}, the time $f(c)$ to process $c$ tokens through a single expert is $f(c) = ac + b$ for $c > 0$, where $b$ represents the weight fetching cost from HBM to on-chip SRAM and $a$ is the marginal computation cost. The total MoE block latency for a sequence of $N$ parallel tokens is thus:
\begin{equation}
\begin{aligned}    
L_{MoE} = \sum_{i=1}^{N_{total}} (b \cdot \mathbbm{1}_{cnt_i > 0} + a \cdot cnt_i) =
b \cdot \left| \bigcup_{n=1}^N \mathcal{S}_n \right| + a \cdot (N \cdot K)
\end{aligned}
\label{eq:latency_model}
\end{equation}
where $cnt_i$ is the number of tokens routed to expert $E_i$, and $\mathcal{S}_n$ is the set of $K$ experts activated by the $n$-th parallel token. 
As a simple illustration, under uniform routing the expected number of unique activated experts $|\bigcup_{n=1}^N \mathcal{S}_n|$ grows as $N_{total}(1 - (1 - K/N_{total})^N)$. Real routers are non-uniform, so measured growth is more gradual (48.1 versus 63.1 experts on LLaDA-MoE 7B at $N=32$), yet it remains well above DES coreset sizes. As the degree of parallelism $N$ increases, this union set expands rapidly (Figure \ref{fig:intro_overview}), leading to an increase in latency (Figure \ref{fig:latency_breakdown}).

This "\textbf{expert explosion}" underscores the limitations of existing expert skipping methods in parallel decoding:
\begin{enumerate}
    \item \textbf{Diminishing Returns from Compute Sparsity:} As latency is dominated by the weight-fetching cost $b \cdot |\bigcup_{n=1}^N \mathcal{S}_n|$, reducing the compute term $a$ yields minimal speedup if an expert remains active for any other concurrent token. This motivates "re-activating" experts within a shared coreset to recover accuracy at near-zero marginal latency (Section~\ref{sec:exp}).
    \item \textbf{Lack of Cross-Token Synergy:} Token-centric skipping ignores the redundancy inherent in parallel decoding. Failing to enforce a collective consensus, they do not minimize the unique expert load $|\bigcup_{n=1}^N \mathcal{S}_n|$, leaving the memory bottleneck unresolved.
\end{enumerate}
Optimizing parallel MoE inference thus requires a shift toward \textit{Dynamic Expert Sharing}, which explicitly minimizes the unique expert load $|\bigcup_{n=1}^N \mathcal{S}_n|$ at the sequence level.

\begin{figure}[t]
    \centering
    \includegraphics[width=\linewidth]{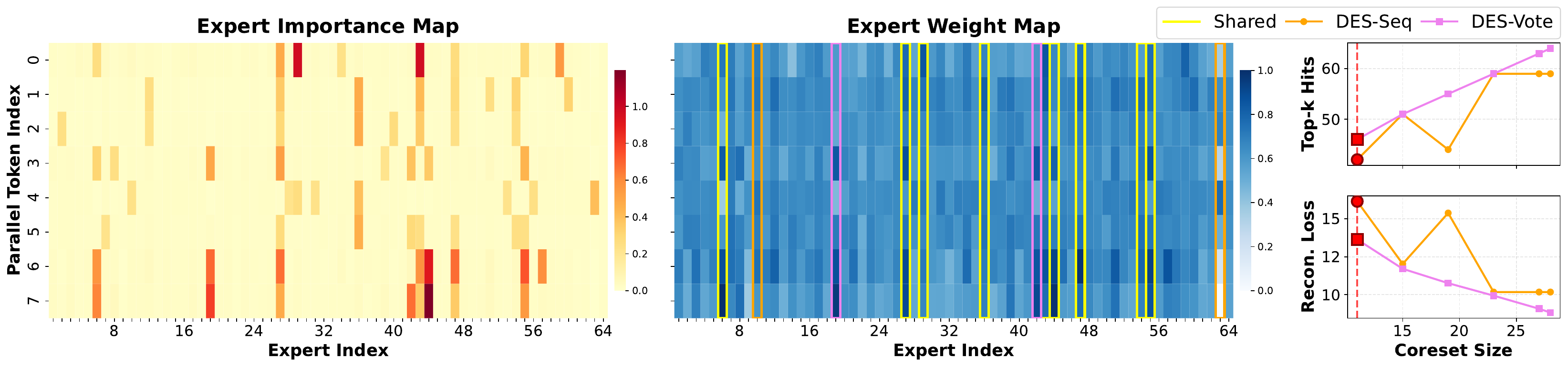}
    \caption{Analysis of Dynamic Expert Sharing (DES) in LLaDA-MoE-7B (8 tokens, Layer 10). (Left) Expert Importance: Heatmap of reconstruction loss sensitivity to expert ablation. Darker regions indicate an increase in reconstruction loss when an expert is removed, revealing strong dependencies between specific token positions and experts.
    (Middle) Selection Overlays: DES-Seq (orange), DES-Vote (pink), and shared (yellow) selections overlaid on log-routing weights. DES-Vote captures high-importance experts (see index 42) that local ranking misses.
    (Right) Performance vs. Coreset Size ($M_{\text{core}}$): DES-Vote achieves higher Top-$k$ recall (top) and lower residual reconstruction loss (bottom) than DES-Seq across $M_{\text{core}}$. Red dashed lines denote the specific $M_{\text{core}}$ in the heatmaps.
    \vspace{-5mm}
    }
    \label{fig:method_expert_analysis}
\end{figure}

\section{Methodology}
\label{sec:method}

\subsection{Dynamic Expert Sharing}

To optimize the memory-bound bottleneck identified in Section~\ref{sec:motivation}, we define a \textbf{Coreset Selection Function}, $\Phi: \mathcal{I} \to \mathcal{C}$, which maps available runtime information $\mathcal{I}$ (e.g., aggregated router logits or hidden states) to a shared subset of experts $\mathcal{C} \subset \{E_1, \dots, E_{M}\}$. 
The coreset selection function identifies the most salient experts for a parallel decoding block. Unlike traditional AR batching, where tokens often belong to disparate tasks, parallel decoding tokens share a common context, a unique opportunity for expert reuse. 
We hypothesize that rather than allowing independent routing, we can identify a compact, sequence-level coreset via cross-token consensus. By restricting each token’s Top-$k$ selection to this shared subset, we maximize expert reuse and minimize HBM transfer.

By substituting this coreset into our latency model, we reformulate the MoE layer latency from Equation~\ref{eq:latency_model}. 
In the vanilla setting, the unique expert load is defined by the union of independent selections $|\bigcup_{n=1}^N \mathcal{S}_n|$, which is implicitly upper-bounded by the total expert pool size $M$. With the introduction of the coreset, the latency is expressed as:
\begin{equation}
L_{\text{MoE}}(\Phi) \le b \cdot |\Phi(\mathcal{I})| + a \cdot (N \cdot k)
\label{eq:coreset_latency}
\end{equation}
where $|\Phi(\mathcal{I})|$ is the cardinality of the selected coreset. Crucially, whereas the vanilla expert load is constrained by a constant, this formulation transforms the upper bound into an explicit variable $|\Phi(\mathcal{I})|$ that guides the optimization process. 
We therefore define the search for an optimal coreset selection strategy as a constrained optimization problem:
\begin{equation}
\begin{aligned}
\Phi^* = \arg \min_{\Phi} \quad & |\Phi(\mathcal{I})| 
\text{s.t.} \quad & \mathcal{A}(\Phi(\mathcal{I})) \geq \mathcal{A}_{\text{base}} - \epsilon 
\end{aligned}
\label{eq:optimization_problem}
\end{equation}
where $\mathcal{A}(\Phi)$ denotes model accuracy and $\epsilon$ is a predefined threshold for tolerable performance degradation. By shifting from independent token-centric routing to collective coreset selection, we decouple the HBM weight-fetching cost from the number of parallel tokens $N$.

The DES algorithm implements this by transitioning from token-level to sequence-level routing. 
As detailed in Algorithm~\ref{alg:des}, the process consists of two stages: (1) \textbf{Sequence-level Consensus}, where the shared coreset $\mathcal{C}$ is identified, and (2) \textbf{Constrained Local Routing}, where each token selects its Top-$k$ experts exclusively from $\mathcal{C}$. 
By re-normalizing weights via an activation function $\sigma$ from the model architecture, we preserve the original routing intent within the optimized memory budget.
The algorithm overview is shown in Figure~\ref{fig:method_overview}.

\subsection{Coreset Selection Methods}

While the optimization problem in Equation~\ref{eq:optimization_problem} is difficult to solve directly, we approximate the solution by proposing two strategies for dynamic coreset selection: \textbf{Intra-Sequence Sharing (DES-Seq)} and \textbf{Saliency-Aware Voting (DES-Vote)}.
These methods serve as the function $\Phi(\mathcal{I})$ in Algorithm~\ref{alg:des} to identify a sequence-wide consensus.

\paragraph{Intra-Sequence Sharing (DES-Seq).} 
A straightforward approach to form a smaller coreset is to select a fixed number of the most salient experts from each token. While previously explored for batch-level optimization in AR models~\cite{oncescu2025opportunistic}, we adapt this to the \textit{intra-sequence} level for dLLM parallel decoding. 
For each token $n$ in the block, we select its top-$k$ experts, where $k$ is a hyperparameter satisfying $k < K$. The coreset $\mathcal{C}$ is the union of these local selections:
\begin{equation}
\mathcal{C}_{\text{DES-Seq}} = \bigcup_{n=1}^{N} \text{TopK}(\mathcal{I}_{n}, k)
\label{eq:des_seq}
\end{equation}
The exact selection algorithm is shown in Algorithm~\ref{alg:des_seq}.

\begin{figure}[t]
\centering
\begin{minipage}[t]{0.5\linewidth}
\vspace{0pt}
\begin{algorithm}[H]
\caption{Dynamic Expert Sharing (DES)}
\label{alg:des}
\small
\begin{algorithmic}[1]
\REQUIRE Sequence info $\mathcal{I}$, Coreset selector $\Phi$, \\Activation $\sigma$, Target $K$.
\STATE \textit{// Stage 1: Sequence-level Consensus}
\STATE $\mathcal{C} \gets \Phi(\mathcal{I})$ \COMMENT{Identify expert coreset}
\STATE \textit{// Stage 2: Constrained Local Routing}
\FOR{each token $n \in \{1, \dots, N\}$}
    \STATE $\mathcal{S}_n \gets \text{TopK}(\mathcal{I}_{n}|_{i \in \mathcal{C}}, K)$ \COMMENT{Route in coreset}
    \STATE $g_n \gets \sigma(\mathcal{I}_{n}|_{i \in \mathcal{S}_n})$ \COMMENT{Re-norm. gates}
    \STATE $y_n \gets \sum_{i \in \mathcal{S}_n} g_{n,i} \cdot E_i(x_n)$
\ENDFOR
\STATE \textbf{return} $Y = \{y_1, \dots, y_N\}$
\end{algorithmic}
\end{algorithm}
\end{minipage}\hfill
\begin{minipage}[t]{0.48\linewidth}
\vspace{0pt}
\begin{algorithm}[H]
\caption{DES-Seq: Intra-Sequence Sharing}
\label{alg:des_seq}
\small
\begin{algorithmic}[1]
\REQUIRE Router logits $\mathcal{I}$, Threshold $k$.
\STATE $\mathcal{C} \gets \bigcup_{n=1}^{N} \text{TopK}(\mathcal{I}_{n}, k)$
\STATE \textbf{return} $\mathcal{C}$
\end{algorithmic}
\end{algorithm}
\vspace{-1.45em}
\begin{algorithm}[H]
\caption{DES-Vote: Saliency-Aware Voting}
\label{alg:des_vote}
\small
\begin{algorithmic}[1]
\REQUIRE Router logits $\mathcal{I}$, Size $M_{\text{core}}$, Top-$K$.
\STATE $\mathcal{I}_{m} \gets \text{Mask}(\mathcal{I}, K)$ \COMMENT{Top-K gates}
\STATE $V \gets \sum_{n=1}^{N} \mathcal{I}_{m, n}$ \COMMENT{Aggregate}
\STATE $\mathcal{C} \gets \text{TopK}(V, M_{\text{core}})$ \COMMENT{Pick coreset}
\STATE \textbf{return} $\mathcal{C}$
\end{algorithmic}
\end{algorithm}
\end{minipage}
\end{figure}

\paragraph{Saliency-Aware Voting (DES-Vote).}
Despite its simplicity, DES-Seq has two primary limitations. First, it does not explicitly maximize expert sharing; it merely reduces local budgets without seeking a global consensus. Second, it utilizes a fixed selection threshold $k$ for all tokens, ignoring that expert importance varies significantly across a sequence (e.g., the 2nd-ranked expert for token A might be more critical than the 2nd-ranked expert for token B).

To address these, we propose \textbf{Saliency-Aware Voting (DES-Vote)}. To maximize consensus, tokens vote for a collective set. 
A naive approach is uniform voting for each token's top-$k$ experts. However, as seen in the \textbf{Expert Importance Map} and \textbf{Expert Weight Map} (Figure~\ref{fig:method_expert_analysis}), there is a high correlation between raw gating weights and actual importance. 
We hypothesize that assigning vote weights based on router scores is effective, as it accounts for varying influences within the top-$k$.

In practice, we aggregate the router weights across the sequence, but crucially \textbf{mask out} weights of experts falling outside each token's local top-$k$ selection to filter noise, and renormalize the remaining weights into each token's top-$k$ gates.
This addresses DES-Seq's second limitation by allowing the collective importance to naturally dictate which experts are retained. The coreset $\mathcal{C}$ is formed by selecting the top $M_{\text{core}}$ experts by total vote:
\begin{equation}
V_i = \sum_{n=1}^{N} \text{Masked}(\mathcal{I}_{n,i}), \quad \mathcal{C}_{\text{DES-Vote}} = \text{TopK}(V, M_{\text{core}})
\label{eq:des_vote}
\end{equation}

As shown in Figure~\ref{fig:method_expert_analysis}, DES-Vote outperforms DES-Seq in both top-$K$ hit rate (preserving more ground truth selections) and reconstruction loss across varying coreset sizes.

\subsection{Custom Kernel for Coreset Selection}\label{subsec:fused_kernel}
To mitigate the system overhead of fragmented operator execution, we develop a custom fused GPU kernel that collapses 12 kernels (e.g., softmax, top-$k$, and reduction) into just two. This design addresses the kernel launch and memory traffic bottlenecks. 
The primary kernel fuses per-token softmax and top-$k$ filtering with weighted expert accumulation, utilizing register-level computation and atomic instructions to update global saliency scores efficiently. 
A second kernel then performs final expert masking based on a threshold-governed ranking.

\subsection{Theoretical Guarantees}
\label{subsec:theory}

We now bound the MoE output perturbation caused by restricting routing to a coreset $\mathcal{C}$, and show that DES-Vote minimizes this bound.
For token $n$, let $\mathcal{S}_n$ be its vanilla top-$K$ set and $\tilde{\mathcal{S}}_n(\mathcal{C})$ its top-$K$ set restricted to $\mathcal{C}$. Let $g_n$ and $\tilde{g}_n(\mathcal{C})$ be the corresponding gate vectors over all $M$ experts, obtained by renormalizing the same nonnegative router scores over $\mathcal{S}_n$ and $\tilde{\mathcal{S}}_n(\mathcal{C})$, and let $y_n$ and $\tilde{y}_n(\mathcal{C})$ be the resulting MoE outputs.
We define the \textbf{discarded routing mass} as $\varepsilon_n(\mathcal{C}) = \sum_{i \in \mathcal{S}_n \setminus \mathcal{C}} g_{n,i}$.
Throughout, we assume $\|E_i(x_n)\| \le B$ for all experts, $|\mathcal{C}| \ge K$, and consistent tie-breaking.

\begin{theorem}[Per-token certificate]
\label{thm:certificate}
For every token $n$, $\|y_n - \tilde{y}_n(\mathcal{C})\| \le 2B\,\varepsilon_n(\mathcal{C})$. In particular, $\varepsilon_n(\mathcal{C}) = 0$ implies $\tilde{y}_n(\mathcal{C}) = y_n$.
\end{theorem}
\begin{proof}
Since $\mathcal{S}_n$ is the global top-$K$ set, every expert in $\mathcal{S}_n \cap \mathcal{C}$ remains in $\tilde{\mathcal{S}}_n(\mathcal{C})$. Replacing excluded experts with lower-scoring ones cannot increase the normalization denominator, hence $\tilde{g}_{n,i}(\mathcal{C}) \ge g_{n,i}$ for all $i \in \mathcal{S}_n \cap \mathcal{C}$. Excluded experts lose exactly $\varepsilon_n(\mathcal{C})$ mass, and as both gate vectors sum to one, surviving and replacement experts gain exactly the same mass, so $\|g_n - \tilde{g}_n(\mathcal{C})\|_1 = 2\varepsilon_n(\mathcal{C})$. The bound follows from $y_n - \tilde{y}_n(\mathcal{C}) = \sum_i (g_{n,i} - \tilde{g}_{n,i}(\mathcal{C})) E_i(x_n)$ and $\|E_i(x_n)\| \le B$. If $\varepsilon_n(\mathcal{C}) = 0$, then $\mathcal{S}_n \subseteq \mathcal{C}$, so $\tilde{\mathcal{S}}_n(\mathcal{C}) = \mathcal{S}_n$.
\end{proof}

\begin{theorem}[Optimality of DES-Vote]
\label{thm:optimality}
Let $V_i = \sum_{n=1}^N g_{n,i}$ be the vote of expert $i$ in Equation~\ref{eq:des_vote}. For any coreset with $|\mathcal{C}| = M_{\text{core}}$, $\sum_{n=1}^N \varepsilon_n(\mathcal{C}) = \sum_{i \notin \mathcal{C}} V_i$. Therefore, $\mathcal{C}_{\text{DES-Vote}}$ minimizes the summed certificate of Theorem~\ref{thm:certificate} over all coresets of size $M_{\text{core}}$.
\end{theorem}
\begin{proof}
As $g_{n,i} = 0$ for $i \notin \mathcal{S}_n$, we have $\varepsilon_n(\mathcal{C}) = \sum_{i \notin \mathcal{C}} g_{n,i}$. Exchanging the sums over $n$ and $i$ gives the identity, which is minimized by keeping the $M_{\text{core}}$ largest votes.
\end{proof}

\begin{corollary}[Arbitrary routing heterogeneity]
\label{cor:worst_case}
Let $V_{[1]} \ge \dots \ge V_{[M]}$ be the sorted votes and $\beta = M_{\text{core}}/M$. For $\mathcal{C} = \mathcal{C}_{\text{DES-Vote}}$,
\begin{equation}
\frac{1}{N} \sum_{n=1}^N \|y_n - \tilde{y}_n(\mathcal{C})\| \le 2B \Big(1 - \frac{1}{N} \sum_{j=1}^{M_{\text{core}}} V_{[j]}\Big) \le 2B(1 - \beta)
\label{eq:worst_case}
\end{equation}
\end{corollary}
\begin{proof}
The first inequality follows from Theorems~\ref{thm:certificate} and~\ref{thm:optimality} with $\sum_i V_i = N$. The second holds as the $M_{\text{core}}$ largest of $M$ nonnegative votes carry at least a $\beta$ fraction of their sum.
\end{proof}

The first bound in Equation~\ref{eq:worst_case} is a data-dependent runtime certificate, and the second holds even when the candidate union $\bigcup_{n=1}^N \mathcal{S}_n$ covers all $M$ experts.

\section{Experiments}
\label{sec:exp}

\subsection{Experimental Setup}

\paragraph{Implementations.}
We evaluate two frontier MoE dLLMs for parallel decoding: the 16B \textbf{LLaDA2.0-mini-preview}~\cite{bie2025llada2} and 7B \textbf{LLaDA-MoE-7B-A1B-Instruct}~\cite{zhu2025llada}. Our experiments utilize the dInfer framework~\cite{ma2025dinfer} with Fast-dLLM~\cite{wu2025fast} as the KV cache method, adopting a 0.9 confidence-based sampling threshold and default hyperparameter configurations. 
All experiments were conducted on \textbf{NVIDIA B200 GPUs}~\cite{nvidia2025blackwell} with CUDA 13.1, paired with an Intel Xeon 6960P CPU. Kernel time was profiled using the NVIDIA Nsight Systems toolkit~\cite{nvidia_nsight_systems}.
For DES, we parameterize DES-Vote using a budget factor $\beta$ such that the coreset size $M_{\text{core}} = \beta \times M$ (where $M$ is the total expert pool size), while DES-Seq is controlled by the local selection count $k$.
We treat $M_{\text{core}}$ as the primary resource budget: results use the same two sizes, $M_{\text{core}} \in \{38, 25\}$, across all tasks and both models (i.e., $\beta \approx 0.15, 0.10$ for the 256 experts of LLaDA2.0-Mini and $\beta \approx 0.60, 0.40$ for the 64 experts of LLaDA-MoE 7B), requiring no per-task or per-dataset search. In deployment, $M_{\text{core}}$ can be set directly from the available expert memory or latency budget; if a specific accuracy target is required, it is a one-dimensional parameter that can be tuned on a small held-out set with efficient methods~\cite{snoek2012practical, li2017hyperband, falkner2018bohb}.

\paragraph{Datasets.}
The algorithm performance is assessed across four benchmarks requiring long-form generative decoding and diverse reasoning: HumanEval~\cite{chen2021codex}, MBPP~\cite{austin2021program}, GSM8K~\cite{cobbe2021gsm8k}, and MATH500~\cite{lightman2023lets}.

\paragraph{Baselines.} 
As existing expert skipping baselines were primarily designed for AR models, we re-implement them for parallel dLLM decoding. Specifically, we adapt \textbf{NAEE}~\cite{lu2024not} by skipping the bottom experts $i$ through $K$ if their cumulative probability $\sum_{u=i}^{K} \pi_{top-u}$ falls below a relative threshold $\beta$ of the total routing sum. We calibrate $\beta$ using the frontier search approach from~\citet{huang2025modes}.
Additionally, we also compare with \textbf{MC-MoE}~\cite{huang2024mixture}, which preserves the experts of the important tokens. 
Notably, we adopt only the dynamic expert skipping components of these baselines, ensuring a fair comparison. 
Finally, we compare against a \textbf{Top-$K$} baseline, which reduces expert traffic by directly decreasing the $K$ value in expert selection.


\begin{table*}[t]
\centering
\caption{
Main results on generative benchmarks for LLaDA2.0-Mini and LLaDA-MoE-7B. 
$\mathcal{T}$ denotes the average number of unique activated experts per layer. 
R.Acc denotes accuracy relative to the \textbf{Vanilla} model. 
\textit{Mem.} is the average memory footprint of unique activated MoE parameters per layer.
All experiments use a block length of 32 with 16 prefix and 16 suffix cache tokens. 
Best results are \textbf{bolded}; the superior performer in adjacent DES-Seq vs. DES-Vote pairs is \underline{underlined}.
}
\label{tab:main_res}
\begin{small}
    \scalebox{0.85}{%
        \begin{tabular}{l | cc cc cc cc | c c c}
\toprule
\multirow{2}{*}{\textbf{Method}} & \multicolumn{2}{c}{\textbf{MBPP}} & \multicolumn{2}{c}{\textbf{GSM8K}} & \multicolumn{2}{c}{\textbf{HumanEval}} & \multicolumn{2}{c|}{\textbf{MATH500}} & \textbf{Avg. $\mathcal{T}$} & \textbf{Mem.} & \textbf{Avg. R. Acc} \\
 & $\mathcal{T}$ & R. Acc. & $\mathcal{T}$ & R. Acc. & $\mathcal{T}$ & R. Acc. & $\mathcal{T}$ & R. Acc. & & GB & \% \\
\midrule
\rowcolor{grey} \multicolumn{12}{c}{\textit{LLaDA2.0-Mini 16B}} \\
\midrule
Vanilla & 78 & 100.0 & 87 & 100.0 & 84 & 100.0 & 85 & 100.0 & 84 & 0.98 & 100.0 \\
\midrule
top-k ($k=4$) & 52 & 70.5 & 58 & 53.8 & 56 & 91.4 & 58 & 83.6 & 56 & 0.66 & 74.8 \\
NAEE ($\beta=0.6$) & 60 & 44.0 & 67 & 61.6 & 67 & 20.5 & 66 & 60.5 & 65 & 0.76 & 46.6 \\
MC-MOE ($\beta=0.6$) & 61 & 45.9 & 67 & 60.2 & 68 & 18.0 & 67 & 62.7 & 66 & 0.77 & 46.7 \\
\cmidrule{2-12}
\rowcolor{highlight} DES-Seq ($k=3$) & 41 & 100.4 & 46 & \textbf{\underline{99.7}} & 45 & 92.4 & 46 & 96.5 & 45 & 0.53 & 97.2 \\
\rowcolor{highlight} DES-Vote ($M_{\text{core}}{=}38$) & \underline{38} & \textbf{\underline{104.7}} & \underline{38} & 98.4 & \underline{38} & \underline{97.5} & \underline{38} & \textbf{\underline{97.4}} & \underline{38} & \underline{0.45} & \textbf{\underline{99.5}} \\
\cmidrule{2-12}
\rowcolor{highlight} DES-Seq ($k=2$) & 31 & 95.1 & 35 & 93.9 & 34 & \textbf{\underline{100.0}} & 35 & 93.6 & 34 & \underline{\textbf{0.40}} & 95.7 \\
\rowcolor{highlight} DES-Vote ($M_{\text{core}}{=}25$) & \textbf{\underline{25}} & \underline{96.4} & \textbf{\underline{25}} & \underline{95.0} & \textbf{\underline{25} }& 97.5 & \textbf{\underline{25}} & \underline{96.5} & \textbf{\underline{25}} & 0.29 & \underline{96.4} \\
\midrule
\rowcolor{grey} \multicolumn{12}{c}{\textit{LLaDA-MoE 7B}} \\
\midrule
Vanilla & 59 & 100.0 & 59 & 100.0 & 59 & 100.0 & 59 & 100.0 & 59 & 0.35 & 100.0 \\
\midrule
top-k ($k=4$) & 46 & 85.6 & 48 & 80.6 & 48 & 79.4 & 48 & 78.9 & 48 & 0.28 & 81.1 \\
NAEE ($\beta=0.6$) & 50 & 96.6 & 50 & 96.6 & 51 & 99.0 & 50 & 93.5 & 50 & 0.29 & 96.4 \\
MC-MOE ($\beta=0.6$) & 52 & 84.8 & 53 & 93.5 & 54 & 79.4 & 53 & 68.2 & 53 & 0.31 & 81.5 \\
\cmidrule{2-12}
\rowcolor{highlight} DES-Seq ($k=3$) & 41 & \textbf{\underline{98.7}} & 42 & 99.1 & 42 & \underline{96.1} & 42 & 93.9 & 42 & 0.25 & 96.9 \\
\rowcolor{highlight} DES-Vote ($M_{\text{core}}{=}38$) & \underline{38} & 97.9 & \underline{38} & \textbf{\underline{100.1}} & \underline{38} & 95.0 & \underline{38} & \textbf{\underline{97.3}} & \underline{38} & \underline{0.22} & \textbf{\underline{97.6}} \\
\cmidrule{2-12}
\rowcolor{highlight} DES-Seq ($k=2$) & 33 & \textbf{\underline{98.5}} & 34 & \underline{98.8} & 34 & 96.1 & 34 & 88.1 & 34 & 0.20 & 95.4 \\
\rowcolor{highlight} DES-Vote ($M_{\text{core}}{=}25$) & \textbf{\underline{25}} & 95.6 & \textbf{\underline{25}} & 98.0 & \textbf{\underline{25}} & \textbf{\underline{101.0}} & \textbf{\underline{25}} & \underline{89.3} & \textbf{\underline{25}} & \textbf{\underline{0.15}} & \underline{96.0} \\
\bottomrule
\end{tabular}
    }
\end{small}
\vspace{-3mm}
\end{table*}

\subsection{Main Results}

As shown in Table~\ref{tab:main_res}, existing expert skipping methods such as \text{NAEE} and \text{MC-MoE} suffer from severe degradation for dLLMs, retaining only $\sim$46\% accuracy on LLaDA2.0-Mini. 
This failure stems from the suboptimality of applying static skipping thresholds across parallel tokens with diverse gating distributions as shown in Figure~\ref{fig:method_expert_analysis}, which often leads to over-sparsification (see Appendix~\ref{appendix:naee} for NAEE calibration).
In contrast, \text{DES} improves performance by exploring dynamic sequence-level expert sharing. 
On LLaDA2.0-Mini, \text{DES-Vote} maintains a 99.5\% relative accuracy compared to vanilla while reducing the unique expert load by 55\%, demonstrating an effective approximation of the optimization objective in Equation~\ref{eq:optimization_problem}. Notably, incorporating local saliency (\text{MC-MoE}) yields negligible improvements over \text{NAEE}, suggesting that token-centric metrics are insufficient to address the collective redundancy inherent in parallel decoding. 
Finally, \text{DES-Vote} outperforms \text{DES-Seq} by achieving higher average accuracy with a smaller expert footprint. 
The relative accuracy of \text{DES-Vote} surpasses that of DES-Seq from 0.6\% to 2.3\%, consistently with lower active experts. These results validate that collective voting of DES-Vote effectively identifies a high-utility coreset that captures expert saliency across the sequence.


\begin{figure}[t] 
    \centering
    \begin{subfigure}{0.495\textwidth}
        \centering
        \includegraphics[width=\linewidth]{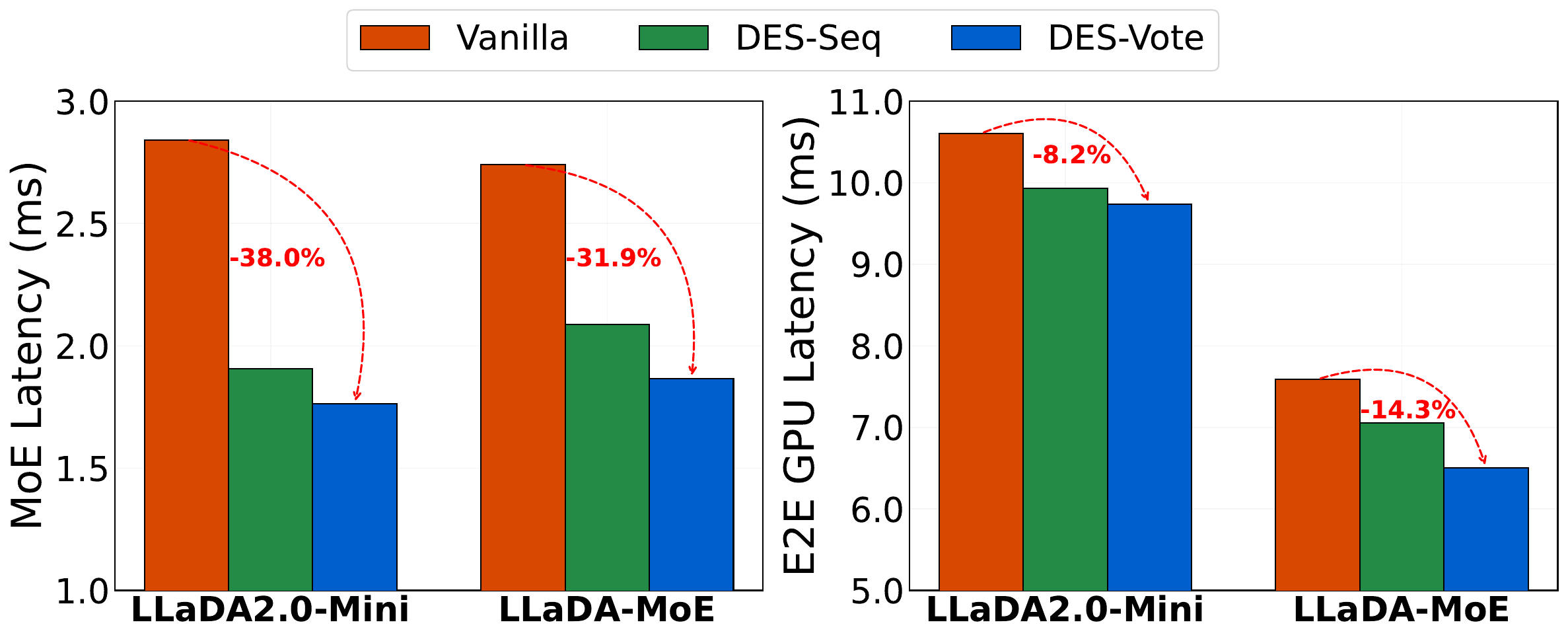}
        \caption{MoE and end-to-end latency across models.}
        \label{fig:latency_comparison}
    \end{subfigure}
    \hfill 
    \begin{subfigure}{0.495\textwidth}
        \centering
        \includegraphics[width=\linewidth]{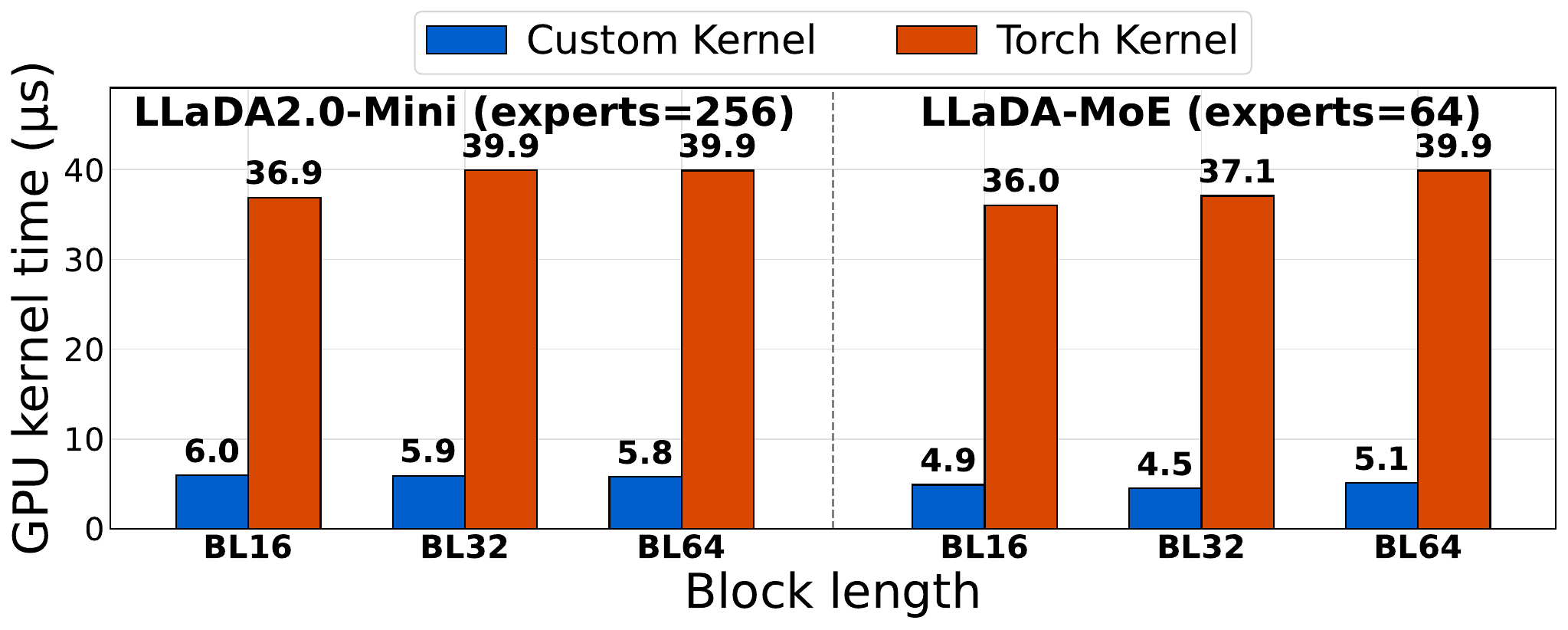}
        \caption{Coreset selection kernel profiling.}
        \label{fig:custom_kernel}
    \end{subfigure}
    \caption{
    \textbf{(a)} Latency measurements for the MoE kernel (left) and total end-to-end GPU kernel execution time (right) across models.
    \textbf{(b)} Profiling of coreset selection latency: our custom fused kernel achieves a large speedup over the PyTorch baseline.
    }
    \label{fig:latency_analysis}
\end{figure}

\subsection{Hardware Efficiency Analysis}

We evaluate the hardware efficiency of \text{DES} on a single NVIDIA B200 GPU. As shown in Figure~\ref{fig:latency_comparison}, \text{DES-Vote} reduces MoE layer latency by up to 38.0\% for LLaDA2.0-Mini and 31.9\% for LLaDA-MoE-7B. Total end-to-end GPU kernel time improves by 8.2--14.3\%. 
The expected dilution in relative gains stems from constant non-MoE operations like self-attention. These results confirm that by mitigating sequence-level HBM traffic, \text{DES} provides substantial wall-clock savings across the entire inference pipeline.
As shown in Figure~\ref{fig:custom_kernel}, our fused kernel~(\secref{subsec:fused_kernel}) achieves a 6$\times$ speedup in coreset selection by eliminating redundant HBM traffic and operator dispatch overhead.

\subsection{Scalability Analysis}
\label{subsec:scalability}

\begin{wraptable}{r}{0.45\textwidth}
\vspace{-4mm}
\centering
\caption{
Batched serving of LLaDA2.0-Mini on a single B200 ($\beta{=}0.10$).
}
\label{tab:batch_results}
\begin{small}
    \scalebox{0.85}{%
        \setlength{\tabcolsep}{2pt}
\begin{tabular}{c | c c c | c c c}
\toprule
\multirow{2}{*}{\textbf{BS}} & \multicolumn{3}{c|}{\textbf{Vanilla}} & \multicolumn{3}{c}{\textbf{DES-Vote}} \\
\cmidrule(lr){2-4} \cmidrule(lr){5-7}
& $\mathcal{T}$ & MoE & E2E & $\mathcal{T}$ & MoE & E2E \\
& & Lat. (ms) & Lat. (ms) & & Lat. (ms) & Lat. (ms) \\
\midrule
\rowcolor{grey} \multicolumn{7}{c}{\textit{Homogeneous (MBPP)}} \\
\midrule
1 & 63.6 & 8.6  & 19.4 & \cellcolor{highlight}25.0 & \cellcolor{highlight}5.8  & \cellcolor{highlight}17.5 \\
2 & 74.1 & 9.4  & 20.7 & \cellcolor{highlight}27.9 & \cellcolor{highlight}5.7  & \cellcolor{highlight}17.6 \\
4 & 77.3 & 10.2 & 23.8 & \cellcolor{highlight}28.4 & \cellcolor{highlight}6.0  & \cellcolor{highlight}19.2 \\
8 & 79.0 & 20.6 & 81.5 & \cellcolor{highlight}30.1 & \cellcolor{highlight}15.2 & \cellcolor{highlight}66.3 \\
\midrule
\rowcolor{grey} \multicolumn{7}{c}{\textit{Heterogeneous (MBPP + GSM8K)}} \\
\midrule
2 & 79.3 & 9.1  & 18.8 & \cellcolor{highlight}27.9 & \cellcolor{highlight}4.6  & \cellcolor{highlight}14.6 \\
4 & 81.1 & 9.9  & 22.4 & \cellcolor{highlight}30.7 & \cellcolor{highlight}5.9  & \cellcolor{highlight}19.3 \\
8 & 83.3 & 16.1 & 60.8 & \cellcolor{highlight}33.0 & \cellcolor{highlight}14.2 & \cellcolor{highlight}56.4 \\
\bottomrule
\end{tabular}

    }
\end{small}
\vspace{-4mm}
\end{wraptable}

\paragraph{Batched serving.}
We extend our analysis to practical online serving that batches multiple requests for higher throughput.
Table~\ref{tab:batch_results} evaluates LLaDA2.0-Mini under homogeneous (MBPP) and heterogeneous (MBPP+GSM8K) batches.
Relative latency gains \emph{grow} with batch size and peak in the moderate range (BS=2--4) reaching up to 41\% MoE-kernel and 23\% E2E latency reduction.
Mixing tasks inflates vanilla expert counts but DES still compresses them effectively.
BS=16 OOMs because dLLMs lack cross-request prefix-cache sharing, placing realistic deployments in the batch size regime where DES delivers the most value.
DES operates along the sequence dimension and is therefore orthogonal to batch-level methods like OEA~\cite{oncescu2025opportunistic}.

\begin{wraptable}{r}{0.45\textwidth}
\vspace{-4mm}
\centering
\caption{
DES on the 103B \textbf{LLaDA2.0-flash} on 4$\times$B200 with tensor parallelism (MBPP).
}
\label{tab:scaling_100b}
\begin{small}
    \scalebox{0.8}{%
        \setlength{\tabcolsep}{4pt}
\begin{tabular}{l c c c }
\toprule
\textbf{Method} & $\mathcal{T}$ & \textbf{R. Acc.}~(\%) & \textbf{MoE Lat.}~(ms)  \\
\midrule
Vanilla & 67.9 & 100.0 & 5.7  \\
\rowcolor{highlight} DES-Vote ($\beta{=}0.10$) & 25.0 & 100.0 & 5.1  \\
\rowcolor{highlight} DES-Vote ($\beta{=}0.05$) & 12.0 & 96.4  & 4.9 \\
\bottomrule
\end{tabular}

    }
\end{small}
\vspace{-4mm}
\end{wraptable}

\paragraph{Scaling up.}
To verify DES at larger scale, we evaluate the 103B \textbf{LLaDA2.0-flash} on MBPP (Table~\ref{tab:scaling_100b}).
DES-Vote ($\beta{=}0.10$) compresses the active footprint by 63\% at no accuracy loss, and a more aggressive $\beta{=}0.05$ shrinks the set to just 12 experts at 96.4\% relative accuracy.
Appendix~\ref{appendix:ep} further studies the performance on expert parallelism (EP).

\begin{figure}[t]
    \centering
    \begin{subfigure}{0.495\textwidth}
        \centering
        \includegraphics[width=\linewidth]{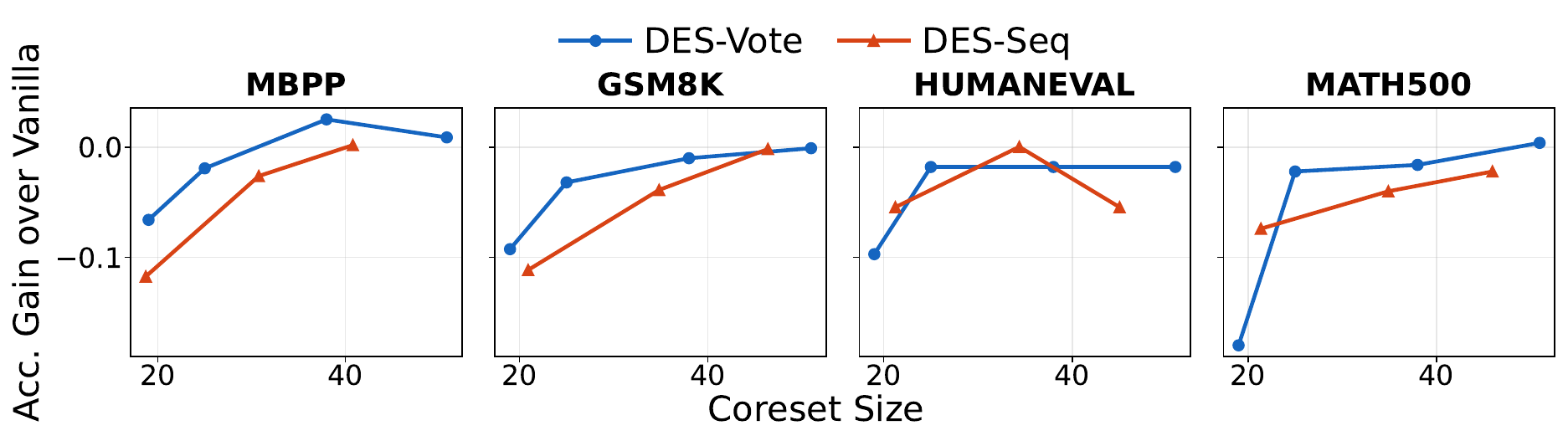}
        \caption{Accuracy gain across varying coreset sizes.}
        \label{fig:acc_delta_coreset}
    \end{subfigure}
    \hfill 
    \begin{subfigure}{0.495\textwidth}
        \centering
        \includegraphics[width=\linewidth]{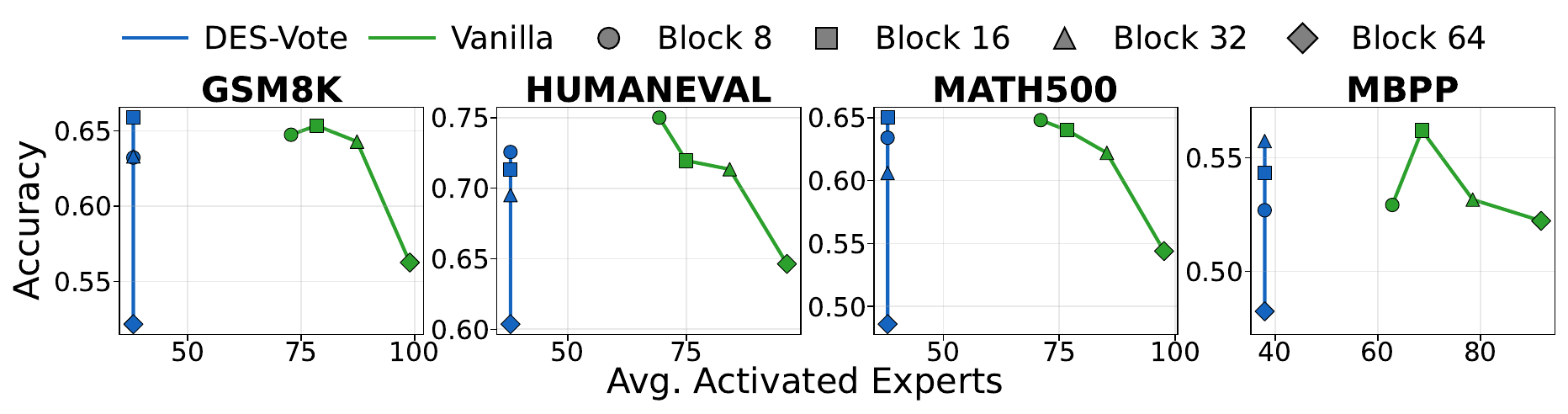}
        \caption{Accuracy vs. active experts across block sizes.
        }
        \label{fig:acc_vs_blocksize}
    \end{subfigure}
    \caption{
    Analysis of efficiency-accuracy trade-offs. All experiments are done with LLaDA2.0-Mini.
    \vspace{-8mm}
    }
    \label{fig:combined_analysis}
\end{figure}

\subsection{Ablation Studies}

\paragraph{Effect of coreset size.}
Figure~\ref{fig:acc_delta_coreset}, model performance generally correlates positively with the coreset size, gradually declining as coreset shrinks. 
Across all benchmarks, \text{DES-Vote} consistently maintains higher accuracy than \text{DES-Seq} at similar coreset sizes. 
Furthermore, by leveraging the continuous $\beta$, \text{DES-Vote} offers greater flexibility in modulating coreset size, enabling extremely small coresets that bypass the one-expert-per-token lower bound inherent to \text{DES-Seq}.

\paragraph{Robustness to block sizes.} Figure~\ref{fig:acc_vs_blocksize} demonstrates the performance of \text{DES-Vote} across varying parallel block sizes ($\{8, 16, 32, 64\}$). The accuracy drop remains consistently small across different tasks as the block size increases, indicating that the coreset selected from \text{DES-Vote}'s collective voting effectively generalizes to larger degrees of parallelism. 
This shows that \text{DES-Vote} maintains a constant and low count of activated experts regardless of block size. In contrast, the vanilla model suffers from a sharp increase in expert activations as the block size grows, leading to a severe memory traffic bottleneck.
This decoupling of memory overhead from parallelism fundamentally shifts the design space for parallel decoding like dLLMs. While recent research emphasizes that block size is vital for balancing algorithmic performance and throughput~\cite{lu2025adablock}, \text{DES-Vote} effectively neutralizes the associated efficiency penalties. 
Consequently, practitioners can optimize block size based purely on the trade-off between multi-token generation efficiency and model accuracy, unconstrained by the traditional limits of memory bandwidth.

\begin{figure}[t] 
    \centering
    \begin{subfigure}{0.495\textwidth}
        \centering
        \includegraphics[width=\linewidth]{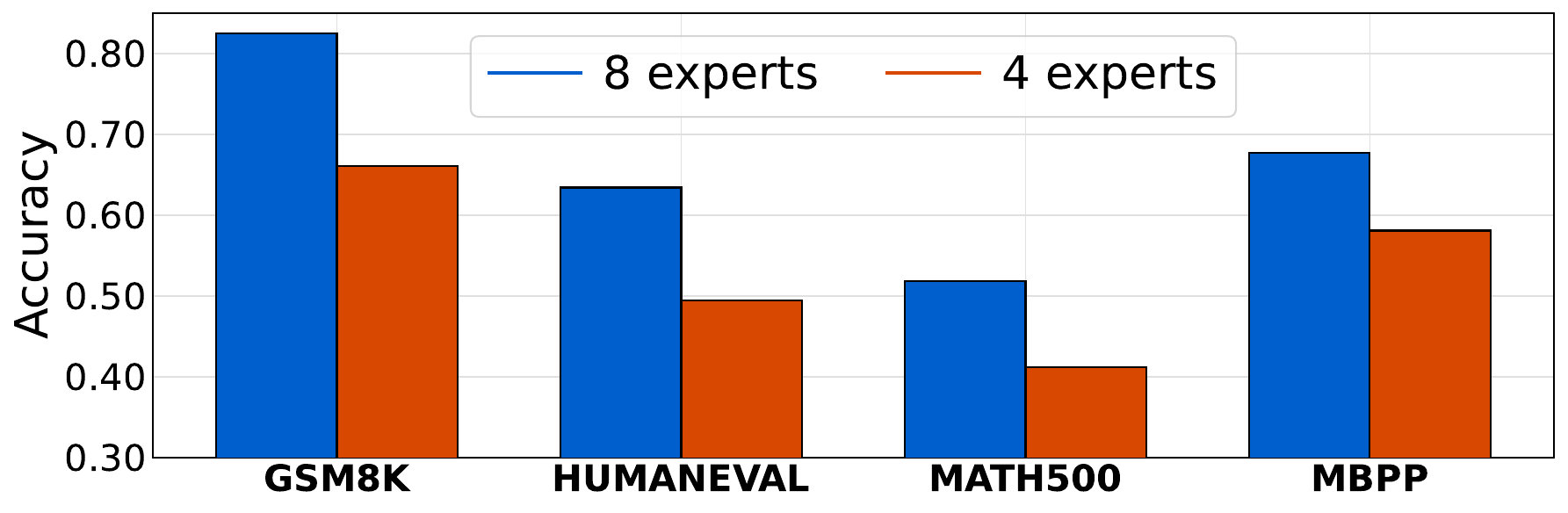}
        \caption{Effect of active expert count on LLaDA2.0-mini.}
        \label{fig:sparsity_ablation}
    \end{subfigure}
    \hfill 
    \begin{subfigure}{0.495\textwidth}
        \centering
        \includegraphics[width=\linewidth]{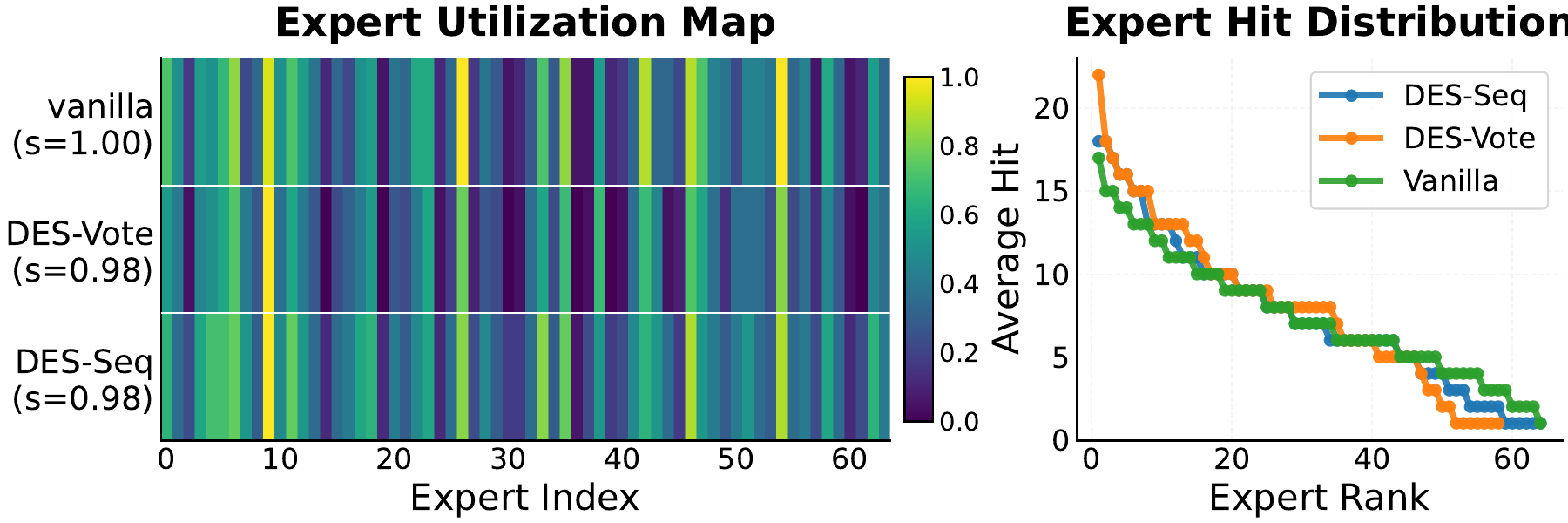}
        \caption{Expert utilization on LLaDA-MoE 7B (MBPP).}
        \label{fig:expert_visualization}
    \end{subfigure}
    \caption{
    \textbf{(a)} Both configurations activate the same number of unique experts, differing only in per-token expert computation ($k=8$ vs. $k=4$). 
    \textbf{(b)} \textbf{(Left)} Normalized hit-rate heatmaps across a parallel block, where $s$ denotes the cosine similarity between the coreset and vanilla expert hit rate vectors. \textbf{(Right)} Expert activation frequencies obtained by ranking experts by their average hit counts within each method. Results are on the 10th layer of LLaDA-MoE 7B.
    \vspace{-5mm}
    }
    \label{fig:expert_analysis}
\end{figure}

\paragraph{Effect of active expert count.} Figure~\ref{fig:sparsity_ablation} compares the performance of DES with different numbers of experts per token. 
Specifically, both configurations identify an equally sized, compact set of experts (Algorithm~\ref{alg:des}, Step 2), hence differing only in their per-token expert computation (Algorithm~\ref{alg:des}, Step 5).
Notably, reducing per-token computation from $8$ to $4$ experts, even while activating the same total number of unique experts, results in a consistent and significant accuracy decrease across all benchmarks. 
This confirms that re-activating experts from the coreset can preserve performance even when these experts differ from the top-8 originally selected by the vanilla model. 


\subsection{Visualization of Expert Utilization}

To evaluate how effectively our methods preserve the model's original routing logic, we analyze expert activation patterns in Figure~\ref{fig:expert_visualization}. First, both \text{DES-Seq} and \text{DES-Vote} exhibit high cosine similarity ($s \geq 0.98$) with the vanilla expert hit rate vector. This high representational fidelity confirms that restricting the expert pool to a coreset does not distort the model's fundamental routing pattern. 
Second, the expert hit distribution reveals a concentration effect. The activation curve for \text{DES-Vote} is noticeably sharper than that of \text{DES-Seq} and the vanilla baseline. \text{DES-Vote} mitigates the long tail of expert hit distribution, hence reducing memory traffic with concentrated expert activation.

\section{Conclusion}
\label{sec:conclusion}

In this paper, we characterize the \textit{expert explosion} phenomenon in dLLM MoE decoding, where the weight-fetching overhead scales linearly with the parallel block size. By identifying a compact, high-utility expert coreset through two strategies, DES-Seq and DES-Vote, our method effectively decouples HBM memory traffic from the degree of parallelism. Extensive experiments demonstrate that DES significantly reduces unique expert activations and MoE latency while retaining accuracy. As memory bandwidth increasingly lags behind processor speed and model expert sparsity grows, we show that DES is a viable approach for realizing the high-throughput potential of parallel generation, motivating further research into sequence-level expert sharing for parallel decoding models.

\textbf{Limitations.} DES is not lossless, and its accuracy--efficiency trade-off varies across tasks and compression levels. The achievable compression and the appropriate budget are model-dependent, and end-to-end gains are smaller than MoE kernel gains since DES does not accelerate attention.

\begin{ack}
This work was conducted during Hao (Mark) Chen's internship at Samsung AI Center, Cambridge, UK.
\end{ack}
\bibliographystyle{plain}
\bibliography{ref}

\newpage
\appendix

\section{Expert Parallelism}
\label{appendix:ep}

While our main experiments isolate algorithmic effects on a single GPU, large-scale MoE deployment commonly partitions experts across devices via Expert Parallelism (EP). EP introduces a new dimension, an irregular all-to-all dispatch of tokens to remote experts, on top of the SRAM-to-HBM weight loading bottleneck.
DES is well-suited to both directions: by enforcing a sequence-level consensus, it (i) directly reduces unique expert traffic in the memory-bound decode regime~\cite{yu2025efficient}, and (ii) shrinks the all-to-all communication volume because fewer distinct experts are addressed in each forward pass.

We empirically verify these benefits using LLaDA2.0-Mini on MBPP under two distributed setups (Table~\ref{tab:ep_results}).
On 2$\times$NVIDIA B6000, DES-Vote cuts the fused MoE kernel latency by 1.76$\times$; on 4$\times$NVIDIA B200, where the per-GPU expert footprint is already small, DES still delivers a 1.10$\times$ MoE-kernel speedup.

\begin{table}[h]
\centering
\caption{Fused MoE kernel latency under EP for LLaDA2.0-Mini on MBPP. DES-Vote uses $\beta=0.10$.}
\label{tab:ep_results}
\begin{small}
    \scalebox{0.95}{%
        \begin{tabular}{l c c | c c}
\toprule
\multirow{2}{*}{\textbf{Method}} & \multicolumn{2}{c|}{\textbf{2$\times$ B6000 (EP=2)}} & \multicolumn{2}{c}{\textbf{4$\times$ B200 (EP=4)}} \\
\cmidrule(lr){2-3} \cmidrule(lr){4-5}
& \textbf{MoE} (ms) & \textbf{Speedup} & \textbf{MoE} (ms) & \textbf{Speedup} \\
\midrule
Vanilla & 5.1 & --- & 1.4 & --- \\
\rowcolor{highlight} DES-Vote & 2.9 & 1.76$\times$ & 1.3 & 1.10$\times$ \\
\bottomrule
\end{tabular}

    }
\end{small}
\end{table}

\section{NAEE Calibration}
\label{appendix:naee}

To verify that the NAEE result on LLaDA2.0-Mini is not caused by one poorly chosen $\beta$, we rerun NAEE on the full MATH500 and MBPP datasets, sweeping flat $\beta$ and reproducing its 25pct and 50pct layerwise calibration protocols (Table~\ref{tab:naee_calibration}).
Calibration improves NAEE only at a substantially larger footprint: the 25pct setting activates 58--72\% more experts than DES-Vote while remaining 9.9--28.0 relative-accuracy points lower. At the tighter 50pct setting, NAEE still activates more experts than DES-Vote but its accuracy collapses. Flat thresholds perform even worse.
This is consistent with their mechanisms: NAEE removes experts independently from each token, so the total number of experts in the block is not explicitly minimized.

\begin{table}[h]
\centering
\caption{NAEE sweep and calibration on LLaDA2.0-Mini. Accuracy is relative to \textbf{Vanilla}.}
\label{tab:naee_calibration}
\begin{small}
    \scalebox{0.95}{%
        \begin{tabular}{l c c | c c}
\toprule
\multirow{2}{*}{\textbf{Setting}} & \multicolumn{2}{c|}{\textbf{MATH500}} & \multicolumn{2}{c}{\textbf{MBPP}} \\
\cmidrule(lr){2-3} \cmidrule(lr){4-5}
& $\mathcal{T}$ & \textbf{R. Acc.}~(\%) & $\mathcal{T}$ & \textbf{R. Acc.}~(\%) \\
\midrule
NAEE, flat $\beta=0.2$ & 69.5 & 39.5 & 63.0 & 15.4 \\
NAEE, flat $\beta=0.35$--$0.70$ & $\le$59.5 & $\le$1.3 & $\le$52.2 & 0.0 \\
NAEE, calibrated 25pct & 65.5 & 87.5 & $\approx$60 & 76.7 \\
NAEE, calibrated 50pct & 52.9 & 4.0 & $\approx$44 & 1.3 \\
\rowcolor{highlight} DES-Vote ($M_{\text{core}}{=}38$) & 38 & 97.4 & 38 & 104.7 \\
\bottomrule
\end{tabular}

    }
\end{small}
\end{table}

\section{Additional Theoretical Remarks}
\label{appendix:theory}

\begin{remark}[Safeguard]
Aggregate optimality in Theorem~\ref{thm:optimality} can still sacrifice a niche token in a highly heterogeneous block. As $\varepsilon_n(\mathcal{C})$ is computable from router outputs before any expert executes, one could enforce $\max_n \varepsilon_n(\mathcal{C}) \le \tau$ by enlarging $\mathcal{C}$ with the highest-weight omitted experts of violating tokens, or by falling back to vanilla routing, guaranteeing $\|y_n - \tilde{y}_n(\mathcal{C})\| \le 2B\tau$ for every token. We do not evaluate this safeguard in Section~\ref{sec:exp}.
\end{remark}

\begin{remark}[Scope]
The bounds in Section~\ref{subsec:theory} are per-layer bounds on MoE output perturbation. A nontrivial bound on discrete benchmark accuracy is impossible without a downstream stability assumption, and composing Lipschitz constants through multiple Transformer blocks makes such a bound very loose. We therefore validate accuracy empirically in Section~\ref{sec:exp}.
\end{remark}

\section{Societal Impact}
\label{appendix:impact}
By mitigating the expert explosion problem, our approach contributes to more energy-efficient and cost-friendly dLLM serving.
There are many potential societal consequences of our work, none of which we feel must be specifically highlighted here.

\newpage
\section*{NeurIPS Paper Checklist}

The checklist is designed to encourage best practices for responsible machine learning research, addressing issues of reproducibility, transparency, research ethics, and societal impact. Do not remove the checklist: {\bf The papers not including the checklist will be desk rejected.} The checklist should follow the references and follow the (optional) supplemental material.  The checklist does NOT count towards the page
limit. 

Please read the checklist guidelines carefully for information on how to answer these questions. For each question in the checklist:
\begin{itemize}
    \item You should answer \answerYes{}, \answerNo{}, or \answerNA{}.
    \item \answerNA{} means either that the question is Not Applicable for that particular paper or the relevant information is Not Available.
    \item Please provide a short (1--2 sentence) justification right after your answer (even for \answerNA). 
\end{itemize}

{\bf The checklist answers are an integral part of your paper submission.} They are visible to the reviewers, area chairs, senior area chairs, and ethics reviewers. You will also be asked to include it (after eventual revisions) with the final version of your paper, and its final version will be published with the paper.

The reviewers of your paper will be asked to use the checklist as one of the factors in their evaluation. While \answerYes{} is generally preferable to \answerNo{}, it is perfectly acceptable to answer \answerNo{} provided a proper justification is given (e.g., error bars are not reported because it would be too computationally expensive'' or ``we were unable to find the license for the dataset we used''). In general, answering \answerNo{} or \answerNA{} is not grounds for rejection. While the questions are phrased in a binary way, we acknowledge that the true answer is often more nuanced, so please just use your best judgment and write a justification to elaborate. All supporting evidence can appear either in the main paper or the supplemental material, provided in appendix. If you answer \answerYes{} to a question, in the justification please point to the section(s) where related material for the question can be found.

IMPORTANT, please:
\begin{itemize}
    \item {\bf Delete this instruction block, but keep the section heading ``NeurIPS Paper Checklist"},
    \item  {\bf Keep the checklist subsection headings, questions/answers and guidelines below.}
    \item {\bf Do not modify the questions and only use the provided macros for your answers}.
\end{itemize}


\begin{enumerate}

\item {\bf Claims}
    \item[] Question: Do the main claims made in the abstract and introduction accurately reflect the paper's contributions and scope?
    \item[] Answer: \answerTODO{} 
    \item[] Justification: \justificationTODO{}

\item {\bf Limitations}
    \item[] Question: Does the paper discuss the limitations of the work performed by the authors?
    \item[] Answer: \answerTODO{} 
    \item[] Justification: \justificationTODO{}

\item {\bf Theory assumptions and proofs}
    \item[] Question: For each theoretical result, does the paper provide the full set of assumptions and a complete (and correct) proof?
    \item[] Answer: \answerTODO{} 
    \item[] Justification: \justificationTODO{}

    \item {\bf Experimental result reproducibility}
    \item[] Question: Does the paper fully disclose all the information needed to reproduce the main experimental results of the paper to the extent that it affects the main claims and/or conclusions of the paper (regardless of whether the code and data are provided or not)?
    \item[] Answer: \answerTODO{} 
    \item[] Justification: \justificationTODO{}

\item {\bf Open access to data and code}
    \item[] Question: Does the paper provide open access to the data and code, with sufficient instructions to faithfully reproduce the main experimental results, as described in supplemental material?
    \item[] Answer: \answerTODO{} 
    \item[] Justification: \justificationTODO{}

\item {\bf Experimental setting/details}
    \item[] Question: Does the paper specify all the training and test details (e.g., data splits, hyperparameters, how they were chosen, type of optimizer) necessary to understand the results?
    \item[] Answer: \answerTODO{} 
    \item[] Justification: \justificationTODO{}

\item {\bf Experiment statistical significance}
    \item[] Question: Does the paper report error bars suitably and correctly defined or other appropriate information about the statistical significance of the experiments?
    \item[] Answer: \answerTODO{} 
    \item[] Justification: \justificationTODO{}

\item {\bf Experiments compute resources}
    \item[] Question: For each experiment, does the paper provide sufficient information on the computer resources (type of compute workers, memory, time of execution) needed to reproduce the experiments?
    \item[] Answer: \answerTODO{} 
    \item[] Justification: \justificationTODO{}
    
\item {\bf Code of ethics}
    \item[] Question: Does the research conducted in the paper conform, in every respect, with the NeurIPS Code of Ethics \url{https://neurips.cc/public/EthicsGuidelines}?
    \item[] Answer: \answerTODO{} 
    \item[] Justification: \justificationTODO{}

\item {\bf Broader impacts}
    \item[] Question: Does the paper discuss both potential positive societal impacts and negative societal impacts of the work performed?
    \item[] Answer: \answerTODO{} 
    \item[] Justification: \justificationTODO{}
    
\item {\bf Safeguards}
    \item[] Question: Does the paper describe safeguards that have been put in place for responsible release of data or models that have a high risk for misuse (e.g., pre-trained language models, image generators, or scraped datasets)?
    \item[] Answer: \answerTODO{} 
    \item[] Justification: \justificationTODO{}

\item {\bf Licenses for existing assets}
    \item[] Question: Are the creators or original owners of assets (e.g., code, data, models), used in the paper, properly credited and are the license and terms of use explicitly mentioned and properly respected?
    \item[] Answer: \answerTODO{} 
    \item[] Justification: \justificationTODO{}

\item {\bf New assets}
    \item[] Question: Are new assets introduced in the paper well documented and is the documentation provided alongside the assets?
    \item[] Answer: \answerTODO{} 
    \item[] Justification: \justificationTODO{}

\item {\bf Crowdsourcing and research with human subjects}
    \item[] Question: For crowdsourcing experiments and research with human subjects, does the paper include the full text of instructions given to participants and screenshots, if applicable, as well as details about compensation (if any)? 
    \item[] Answer: \answerTODO{} 
    \item[] Justification: \justificationTODO{}

\item {\bf Institutional review board (IRB) approvals or equivalent for research with human subjects}
    \item[] Question: Does the paper describe potential risks incurred by study participants, whether such risks were disclosed to the subjects, and whether Institutional Review Board (IRB) approvals (or an equivalent approval/review based on the requirements of your country or institution) were obtained?
    \item[] Answer: \answerTODO{} 
    \item[] Justification: \justificationTODO{}

\item {\bf Declaration of LLM usage}
    \item[] Question: Does the paper describe the usage of LLMs if it is an important, original, or non-standard component of the core methods in this research? Note that if the LLM is used only for writing, editing, or formatting purposes and does \emph{not} impact the core methodology, scientific rigor, or originality of the research, declaration is not required.
    \item[] Answer: \answerTODO{} 
    \item[] Justification: \justificationTODO{}

\end{enumerate}

\end{document}